\documentclass[runningheads]{llncs}
\usepackage{graphicx}
\usepackage{comment}
\usepackage{amsmath,amssymb} 
\usepackage[ruled,vlined]{algorithm2e}

\usepackage[lofdepth,lotdepth]{subfig}
\usepackage{epsfig}
\usepackage{graphicx}
\usepackage[usenames, dvipsnames]{color} %
\usepackage{xcolor}
\usepackage[normalem]{ulem}
\newcommand\mycommfont[1]{\footnotesize\ttfamily\textcolor{blue}{#1}}
\SetCommentSty{mycommfont}

\usepackage{booktabs}
\usepackage{multirow}
\usepackage{url}

\begin{document}

\pagestyle{headings}
\mainmatter
\def\ECCVSubNumber{6425}

\title{Online Continual Learning under Extreme Memory Constraints} 

\titlerunning{Online Continual Learning under Extreme Memory Constraints}

\author{Enrico Fini\inst{1, 2} \and 
St\'{e}phane Lathuili\`{e}re\inst{3}
\and
Enver Sangineto\inst{1} \and
Moin Nabi\inst{2} \and \\
Elisa Ricci\inst{1,4}}
\authorrunning{E. Fini et al.}
\institute{University of Trento, Trento, Italy \and
SAP AI Research, Berlin, Germany \and
LTCI, T\'{e}l\'{e}com Paris, Institut Polytechnique de Paris, Palaiseau, France \and
Fondazione Bruno Kessler, Trento, Italy \\
\email{enrico.fini@unitn.it}}

\maketitle

\begin{abstract}
Continual Learning (CL) aims to develop agents emulating the human ability to sequentially learn new tasks while being able to retain knowledge obtained from past experiences. In this paper, we introduce the novel problem of Memory-Constrained Online Continual Learning (MC-OCL) which imposes strict constraints on the
memory overhead that a possible algorithm can use to avoid catastrophic forgetting. As most, if not all, previous CL methods violate these constraints, we propose an algorithmic solution to MC-OCL: Batch-level Distillation (BLD), a regularization-based CL approach, which effectively balances stability and plasticity in order to learn from data streams, while preserving the ability to solve old tasks through distillation. Our extensive experimental evaluation, conducted on three publicly available benchmarks, empirically demonstrates that our approach successfully addresses the MC-OCL problem and achieves comparable accuracy to prior distillation methods requiring higher memory overhead.
\end{abstract}

\section{Introduction}
\label{Introduction}

A well-known problem in deep learning is the tendency of Deep Neural  Networks (DNNs)
to {\em catastrophically forget} the knowledge acquired from old tasks when learning a new task. Differently from humans, who have the natural ability to selectively retain knowledge obtained through past experience when facing a new problem or task, a DNN, trained on a given data distribution, tends to be drastically affected 
when new training data drawn from a different distribution are provided, losing the ability to solve the past task(s). 
{\em Continual Learning} (CL) \cite{DeLange-CL-survey} investigates this {\em stability-plasticity dilemma}: how can a DNN be adapted to solve a new task without losing the ability to deal  with previously seen tasks?

\begin{figure}[t]
\centering
\includegraphics[width=0.95\textwidth]{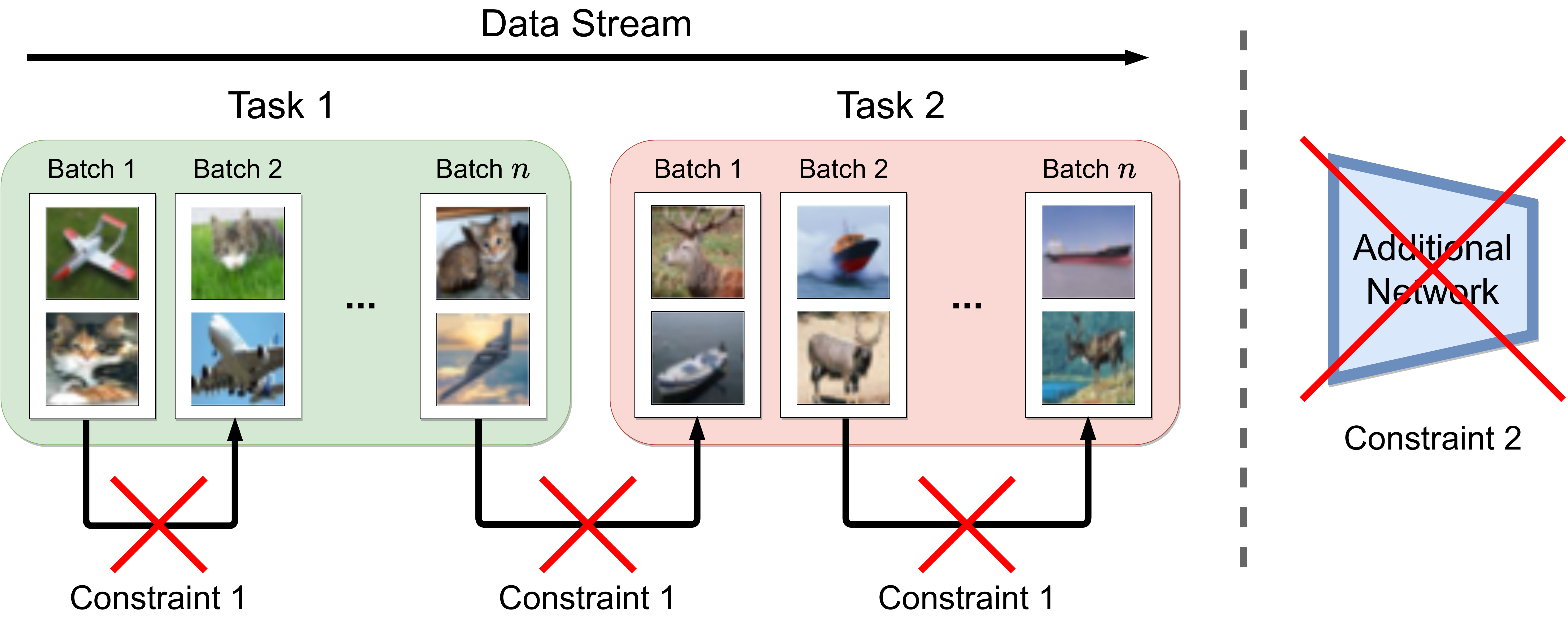}
\caption{Illustration of the proposed Memory-Constrained Online Continual Learning setting, where two constraints should be satisfied: (1) No information should be transferred between data batches and between tasks; (2) No memory can be allocated for auxiliary networks or network expansions.}
\label{fig:sheme}
\end{figure}

Due to the relevance of its applications, in the last few years, the computer vision research community has put considerable effort into developing CL methods.
Previous work in the field can be categorized according to the strategy used to mitigate catastrophic forgetting \cite{DeLange-CL-survey}. Replay-based methods  \cite{rebuffi2017icarl,castro2018end,hou2019learning,wu2019large}, for instance, alleviate forgetting by storing old data or synthesizing virtual samples from the past. Parameter-isolation  approaches \cite{mallya2018packnet,rusu2016progressive} dedicate  specific portions of the network parameters to each task. Finally, regularization-based methods \cite{li2017learning,dhar2019learning,kirkpatrick2017overcoming,aljundi2018memory} introduce additional regularization terms in the loss function to encourage the stability of the network with respect to the previous tasks. 
 Specifically, regularization may be obtained using a distillation-like \cite{hinton2015distilling} approach  or enforcing a prior on the model parameters.
 In the first case,  the network is encouraged to keep the predictions consistent with respect to the old tasks \cite{li2017learning}.
 Prior-based methods, on the other hand, estimate and store a prior on the parameter distribution which indicates the importance  of each parameter with respect to the old tasks
 \cite{kirkpatrick2017overcoming}.

{\em Online Learning} (OL) studies optimization methods which can operate with a {stream} of data: learning goes on as the data are collected
\cite{DeLange-CL-survey,Duda2000}. A typical application  of OL are those scenarios in which training data cannot be stored (e.g., due to memory restrictions or data privacy concern). While classic OL assumes i.i.d. data sampling over a single task, in this paper we deal with {\em Online Continual Learning} (OCL), where data are provided with a sequential stream and the data distribution undergoes drastic changes when a new task is introduced. Previous works in this field \cite{aljundi2019task,aljundi2019gradient} mainly focus on the {task-free} scenario, in which no task-boundary information is provided. However, the solutions they propose rely on either a buffer or a generator to replay data from previous time steps. On the one hand, the buffer-based solution violates a strict online regime, where training data from past time steps should be discarded. On the other hand, a generator network involves a big memory and computational overhead that needs to be allocated on purpose. 

Conversely, in this paper, we introduce a novel problem, Memory-Constrained Online Continual Learning (MC-OCL), where we impose strict memory constraints during the course of training. Specifically, we want to minimize the memory overhead, while preserving the utility of the network. 
This implies the network to discard all the unnecessary information for inference. Specifically, we argue that a memory-efficient OCL approach should satisfy the two conditions (see Fig.\ref{fig:sheme}): (1) No information should be passed from a generic time step $s$ to time step $s+1$, except the network itself; (2) No memory can be allocated for network expansions or dedicated as auxiliary networks. Note that constraint (1) does not only imply that each batch is treated independently but also excludes information pass through subsequent tasks. The proposed constraints are  particularly relevant for those application scenarios in which the network is deployed on devices with small memory footprint (e.g. robots or smartphones) or in which past images cannot be stored due to privacy issues.

Currently, existing CL solutions cannot deal with the proposed MC-OCL scenario. In fact, replay-based methods \cite{rebuffi2017icarl,hou2019learning,ostapenko2019learning} need to either explicitly store (part of the) training samples (violating constraint (1)) or to train an ad-hoc generator network (violating constraint (2)). Even regularization-based methods using distillation \cite{li2017learning,dhar2019learning} either need to store the model output probabilities (violating constraint (1)) or task-specific networks (violating constraint (2)) in order to produce distillation information on the fly. Finally, parameter-isolation based methods \cite{mallya2018packnet,rusu2016progressive} which select a subset of the network parameters for each task, assume that task-specific information (e.g. mask in \cite{mallya2018packnet}) can be transferred from different tasks and do not perform data stream processing on a mini-batch basis (violating constraint (1)).

In this paper, we propose a conceptually simple yet empirically powerful solution to the MC-OCL problem called Batch-level Distillation (\textit{BLD}), in which distillation information is re-generated at each time step without violating constraint (1). Our approach is articulated into two main phases. In the first stage, the {\em warm-up}, data of the current batch are exploited to perform a first gradient descent step minimizing the cross-entropy loss over the new task classifier. The predictions of the old task classifiers are stored in a probability bank that is required in the second stage, referred to as {\em joint training}. In this stage, both the distillation and the new task learning are performed, adopting a dynamic weighting strategy that uses the gradient norm computed in the {\em warm-up} stage. We extensively evaluate the proposed solution on three widely-used benchmarks: {MNIST} \cite{lecunmnist}, {SVHN} \cite{netzer2011reading} and {CIFAR10} \cite{cifar}. Our results demonstrate that \textit{BLD} achieves comparable accuracy to state of the art distillation methods despite the imposed memory constraints.

To summarize, our contributions are the following:
\begin{itemize} \renewcommand{\labelitemi}{$\bullet$} 
    \item We introduce a realistic yet challenging OCL setting which operates under extreme memory constraints (MC-OCL).
    \item We propose the notion of Batch-level Distillation (\textit{BLD}) as a viable solution to the MC-OCL problem.
    \item An extensive empirical study is carried out which confirms the effective alleviation of forgetting despite the strict memory constraints.
\end{itemize}

\section{Related work}

Over the past few years, Continual Learning \cite{DeLange-CL-survey,kirkpatrick2017overcoming} has received increased interest in computer vision. Indeed, CL is highly relevant for several applications. For instance, for object recognition it is very desirable to dispose of deep models which are able to recognize new object classes, while retaining their
knowledge on the categories they have been originally trained for.
Previous CL methods can be roughly categorized into three main groups \cite{DeLange-CL-survey}:  regularization-based \cite{li2017learning,dhar2019learning,kirkpatrick2017overcoming,aljundi2018memory}, parameter-isolation based  \cite{mallya2018packnet,rusu2016progressive} and replay-based \cite{rebuffi2017icarl,hou2019learning,ostapenko2019learning} methods. 

Data-focused regularization-based methods \cite{li2017learning,dhar2019learning} develop from the idea of applying the distillation paradigm \cite{hinton2015distilling} to prevent catastrophic forgetting. One of the earlier approaches in this category is Learning without Forgetting (LwF) \cite{li2017learning}, where a distillation loss is introduced to preserve information of the original classes considering the output probabilities. LwF exploits data from the original classes during training when the classifier is trained to recognize novel categories. Recently, the concept of distillation has been extended to attention and segmentation maps~\cite{dhar2019learning,cermelli2020modeling}. 

Prior-focused regularization-based methods \cite{kirkpatrick2017overcoming,aljundi2018memory}
consider the network parameter values as a source of knowledge to be transferred and operate by penalizing changes of parameters relevant for old tasks when learning on the new task. These approaches mostly differ in the way parameter relevance is computed. 
A prominent work in this category is Elastic Weight Consolidation (EWC) \cite{kirkpatrick2017overcoming}, where parameter update rules are obtained approximating the posterior as a Gaussian distribution. Differently, Aljundi \textit{et al.} propose Memory Aware Synapses (MAS) \cite{aljundi2018memory}, an approach that estimates the network weight importance using small perturbations of the parameters.

Parameter-isolation based approaches \cite{mallya2018packnet,rusu2016progressive} address catastrophic forgetting by allocating specific model parameters to each task. For instance, in \cite{mallya2018packnet} a fixed architecture is considered and parts that are specific for some previous tasks are masked out while training on novel tasks. Rusu \textit{et al.} \cite{rusu2016progressive} proposed Progressive Neural Networks (PNNs), a framework which transfers across sequences of tasks by retaining a pool of pre-trained models and learning connections in order to get useful features for the novel task.

Replay-based methods alleviate catastrophic forgetting by either storing  \cite{rebuffi2017icarl,castro2018end,hou2019learning,wu2019large} or by artificially generating \cite{ostapenko2019learning,shin2017continual} images of previous tasks, often referred to as exemplars. Based on this idea Rebuffi \textit{et al.} propose ICARL where a strategy to select exemplars in combination with a distillation loss is introduced. Subsequent works \cite{chaudhry2018riemannian,wu2019large} further analyze exemplars selection strategies. Differently, other works \cite{ostapenko2019learning,shin2017continual,wu2018memory} propose to employ generative networks to generate synthetic data of old tasks. However, these methods significantly depend on the network capacity and struggle to generate high-quality images.

Our approach belongs to the category of data-focused regularization-based methods, as it also attempts to counteract catastrophic forgetting through distillation. However, differently from previous methods we focus on an online setting where no information is passed through different tasks and batches.

Recently, few works in CL have considered an online CL setting \cite{aljundi2019gradient,lee2020neural,aljundi2019task,aljundi2019online}. However, they mostly focus on task-free continual learning, developing methodologies to automatically detect task boundaries and address the online learning problem benefiting from specific buffers. Our work develops with a different perspective as we aim to design an OCL framework maintaining memory requirements at minimum, thus assuming that no information is retained when processing the next batch in the data stream. 

Finally, MER \cite{riemer2018learning} and OML \cite{javed2019meta} are two recent meta-learning approaches to continual learning. However, the former needs a very large buffer (1k samples per task). On the other hand, OML, does not require any buffer, but works with very short tasks, while we use much larger datasets. Also, OML is based on an offline meta-pretraining, while we train the whole network from scratch.

\section{Memory-Constrained Online Continual Learning}
\label{Setting}

\subsection{Problem and Notation}
\begin{figure}[t]
\centering
\includegraphics[width=0.90\textwidth]{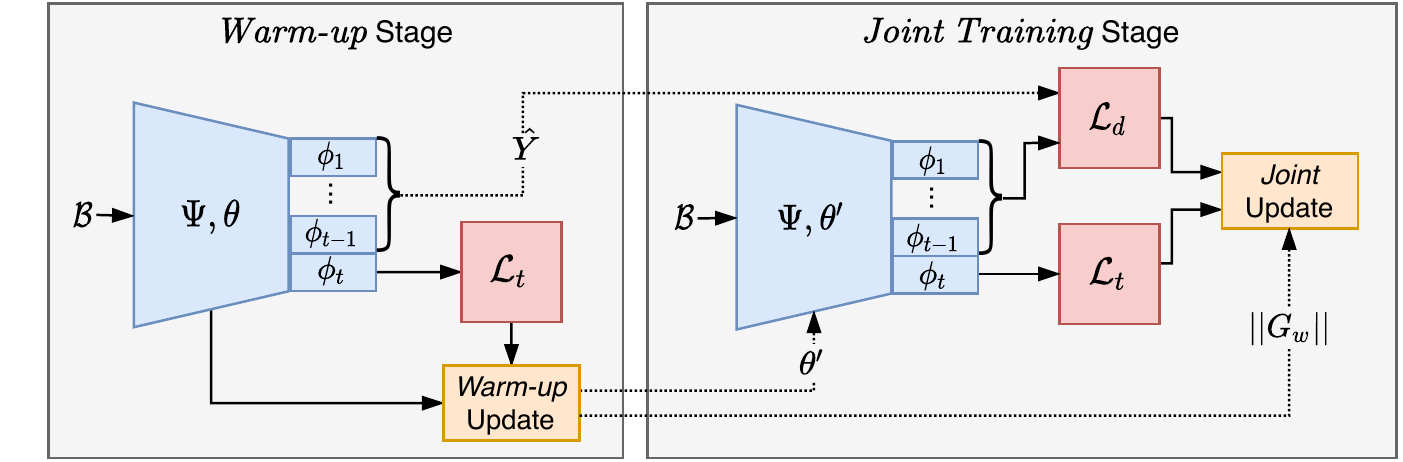}
\caption{Overview of \textit{BLD}: considering the current batch $\mathcal{B}$, we proceed in two stages. In the \emph{warm-up} stage, we perform a first gradient descent step minimizing the cross-entropy loss over the new task classifier. The predictions of the old task classifiers are stored in a probability bank. The \emph{joint training} stage performs knowledge distillation to prevent forgetting, and new task learning employing a dynamic weighting strategy that uses the gradient norm computed in the \emph{warm-up} stage.}
\label{fig:pipeline}
\end{figure}

Without loss of generality, a typical CL scenario can be formalized assuming a set  $\mathcal{T} = \{ T_1, ..., T_n \}$ of $n$ different tasks, where each task is characterized by a different joint probability distribution $P_t$ of the raw images $x \in \mathcal{X}_t$ and the class labels $y \in \mathcal{Y}_t$. During time $t$, a new task $T_t$ is presented to the DNN (see Fig.~\ref{fig:pipeline}) and the goal is to learn $T_t$ without catastrophically forgetting  $T_1, ... T_{t-1}$. Note that not only the set of images $\mathcal{X}_t$ is task-specific, but so is the corresponding set of possible labels $\mathcal{Y}_t$. Following common practice in CL literature, we assume that the task-change event is known, and when a new task $T_{t+1}$ arrives we ask the network to learn to classify the new images according to $\mathcal{Y}_{t+1}$, being simultaneously able to solve the old tasks. 

In this paper, we assume that our classification network is composed of a backbone, the {feature extractor} $\Psi$, and multiple heads $\Phi = \{\phi_1, \hdots, \phi_n\}$, where the $t$-th head $\phi_t$ is composed of a linear classifier with a softmax activation which computes task-specific classification probabilities over $\mathcal{Y}_t$. In addition, $\phi_t$ also accepts an optional temperature parameter $\tau$.

In the proposed MC-OCL setting the memory overhead must be kept at minimum. To fulfill this requirement we set several constraints. We impose that, when learning a new task ${T}_t$, the only memory overhead are the parameters of each task-specific classifier $\phi_t$, while $\Psi$ is shared over all the tasks and no other high-capacity network can be used to solve the CL problem (constraint (2)).
In addition, it is reasonable to suppose that the complete dataset of the task cannot fit in memory. Consequently, standard batch training procedures consisting in observing several times each sample cannot be applied. Training must be addressed following an online formulation. More precisely, we assume that only a mini-batch of data $\mathcal{B}$ associated with task $T_t$ is available at every time step. Importantly $\mathcal{B}$ contains only a few data (e.g., a few dozen images). This ``mini-batch'' based relaxation of the typical OL scenario \cite{Duda2000} is commonly adopted in other COL settings \cite{aljundi2019online}.
Moreover, in our MC-OCL setting, every information, except the network parameters, must be discarded after processing each batch $\mathcal{B}$ (constraint (1)). $\mathcal{B}$ is used to update the network weights, but no explicit information can be stored or passed to the next batch processing step

\subsection{Batch-level Distillation}
\label{sec:method}

In this section, we describe the proposed method, named Batch-level Distillation (\textit{BLD}). Inspired by \cite{li2017learning}, we adopt a formulation based on knowledge distillation to mitigate catastrophic forgetting.
Our distillation approach is composed of two main stages, both depending only on the current mini-batch data $\mathcal{B}$ which is sampled from the data distribution of the current task $T_t$ and on the network parameters $\theta$. The overall pipeline is illustrated in Fig.~\ref{fig:pipeline}. The first stage, named {\em warm-up} stage, is introduced in order to enable the use of knowledge distillation in the second stage, named {\em joint training} stage. 

The key idea of distillation for CL, is to use a regularization loss which prevents  that the predictions of the old task classifiers are significantly modified  when learning the new task. Since we only have available a mini-batch $\mathcal{B}$ of data, we propose to apply the distillation paradigm at the mini-batch level rather than at the dataset level. In other words, we  enforce that, while learning the new task, the predictions of the old classifiers do not change much {\em between the beginning and the end of the current mini-batch processing}. This regularization and the new-task loss are optimized together in the {\em joint training} stage.

In order to use a distillation regularization, we need to estimate the predictions of the old task classifiers before updating the network parameters. This is the main purpose of the \emph{warm-up} stage. In addition to computing the old task predictions, the \emph{warm-up} stage also performs a first learning step by minimizing the new task  loss. As detailed in Sec.~\ref{sec:joint}, this initial learning step is required in order to perform distillation in the second stage. Finally, the \emph{warm-up} stage is also used to estimate the gradient norm that is later used in the second stage to obtain a dynamic weighting of the different loss terms. 
We now provide the details of the two stages. 

\subsection{Warm-up Stage}
The purpose of this first stage is threefold: collecting distillation data (used only in the second stage), starting learning the new task  on the current batch and estimate the norm of the new task loss.
The details of the warm-up stage are provided in Alg.\ref{alg:warm}.
\SetKwFor{For}{for}{}{end for}%

\begin{algorithm}[H]
\DontPrintSemicolon
\label{alg:warm}
  \SetAlgoLined
  \SetKwInOut{Input}{Input}
  \SetKwInOut{Output}{Output}
  \SetKwInOut{return}{Return}
  \Input{Current network ($\Psi$, $\Phi$, $\theta$), current batch $\mathcal{B}$ with labels $\mathbf{y}$, learning rate $\alpha_{w}$, temperature $\tau$}
     $\mathbf{v} = \Psi(\mathcal{B}; \theta)$ \tcp*{feature extraction}
    $\hat{Y} = \varnothing$ \tcp*{initialize empty probability bank}
    \For(\mycommfont{\hfill // for every past task}){$o \in \{ 1, ..., t-1 \}$}{
      $\hat{\mathbf{y}}^o=\phi_o(\mathbf{v},\tau;\theta)$ \tcp*{compute predictions}
      $\hat{Y}\leftarrow \hat{Y}\bigcup\{\hat{\mathbf{y}}^o\}$  \tcp*{fill probability bank}
    }
    $\mathcal{L}_{t}=H(\phi_{t}(\mathbf{v};\theta),\mathbf{y})$  \tcp*{compute warm-up loss}
    $G_w = \frac{\partial \mathcal{L}_{t}}{\partial \theta}$ \tcp*{compute warm-up gradient}
    $\theta'=\theta-\alpha_{w}G_w$ \tcp*{parameter update}
    \Return{$\theta', \hat{Y}, ||G_w||$}
 \caption{{\em Warm-up} Stage}
\end{algorithm}

Specifically, let $\theta$ be the set of all the parameter values in $\Psi$ and $\Phi$. 
Considering an image $x_b \in \mathcal{B}$, we use the current feature extractor $\Psi$ to get $\mathbf{v}_b = \Psi(x_b; \theta)$. We introduce the notation $\mathbf{v}=\{\mathbf{v}_b, 1\leq b\leq|\mathcal{B}|\}$ to indicate all the images of the current batch, and we simply write $\mathbf{v} = \Psi(\mathcal{B}; \theta)$.

Then, we use these features to compute the predictions for the new images using the old task classifiers. More specifically, for each old task $T_o$, we estimate $\hat{\mathbf{y}}^o =\{\hat{\mathbf{y}}_b^o,1\leq b\leq|\mathcal{B}|\} = \phi_o(\mathbf{v,\tau;\theta})$, where $\tau$ is the temperature of the softmax. These probability vectors are then appended to a probability bank $\hat{Y}$. At the end of the \textit{warm-up} stage, $\hat{Y}$ will contain the predicted probabilities for every image of the batch according to every old classifier. This memory is later used for distillation in the second stage but it is released  before receiving the next data batch. Since the number of classes is relatively small
(hence, each $\hat{\mathbf{y}}_b^o$ is a low-dimensional vector), 
the memory required to store $\hat{Y}$ remains negligible compared to the memory space used by the batch of input images and the network parameters.

The previously computed features $\mathbf{v}$ are used also by the  new-task classifier for computing the standard cross-entropy loss. Specifically, given the features $\mathbf{v}$ and their corresponding one-hot labels $\mathbf{y} \in \{0,1\}^{|\mathcal{B}| \times |\mathcal{Y}_t|}$, we use:
\begin{equation}
\label{eq.cross-entropy-loss}
\mathcal{L}_t =H\left(\phi_{t}(\mathbf{v};\theta),\mathbf{y}\right) = - \sum^{|\mathcal{B}|}_{b=1} \mathbf{y}_b\cdot\log \phi_{t}\left(\mathbf{v}_b;\theta\right).
\end{equation}
Then, the gradient $G_w = \frac{\partial \mathcal{L}_{t}}{\partial \theta}$ is computed, and the parameters of the network are updated using the standard gradient descent. The warm-up stage also returns the norm of the gradient $||G_w||$, which is used  for the parameter normalization in the second stage ({\em joint training} stage). In practice, since the norm of the gradient is computed layer-wise (see later), it can be obtained during the backward pass without  storing the gradient of the whole network.

To conclude this stage, the memory used by the intermediate variables (e.g. $\mathbf{v}$ and $\theta$) is released. At this point, the memory contains the parameters $\theta'$, the batch $\mathcal{B}$, the probability bank $\hat{Y}$ and the norm of the gradient $||G_w||$.

\subsection{Joint training stage}
\label{sec:joint}
We now provide the description of the joint training stage. The goal of this stage is to update the network parameters with respect to the new task while preserving the knowledge of the previous tasks. The details are provided in Alg.\ref{alg:joint}.

\begin{algorithm}[H]
\DontPrintSemicolon
\label{alg:joint}
  \SetAlgoLined
  \SetKwInOut{Input}{Input}
  \SetKwInOut{Output}{Output}
  \SetKwInOut{Return}{Return}
  \Input{Current network ($\Psi$, $\Phi$, $\theta'$), current batch $\mathcal{B}$ with labels $\mathbf{y}$, old task probability bank $\hat{Y}$, learning rate $\alpha_{j}$, temperature $\tau$, gradient-balancing factor $\lambda$, norms of the gradients $||G_w||$.}
    $\mathbf{v}' = \Psi(\mathcal{B}; \theta')$ \tcp*{feature extraction}
    
    $\mathcal{L}_{d} = \sum_{o=1}^{t-1}H(\phi_{o}(\mathbf{v}',\tau;\theta'),\hat{\mathbf{y}}^o)$ \tcp*{compute distillation loss}
    $G_j =\frac{\partial \mathcal{L}_{d}}{\partial \theta'}$  \tcp*{distillation gradient}

    $G_j \gets \lambda\frac{||G_w||}{||G_j||} \, G_j $ \tcp*{balance the distillation gradient}

    $\mathcal{L}_{t} = H(\phi_{t}(\mathbf{v}';\theta'),\mathbf{y})$ \tcp*{compute new task loss}
    
    $G_j \leftarrow G_j + \frac{\partial \mathcal{L}_{t}}{\partial \theta'}$ \tcp*{accumulate new task gradient}
    
    $\theta'' = \theta'-\alpha_{j}G_j$ \tcp*{parameter update}
\Return{$\theta''$}
 
\caption{{\em Joint Training} Stage}
\end{algorithm}
    
    


    
    


Using the current batch $\mathcal{B}$ and the {\em updated} feature extractor $\Psi(\cdot ; \theta')$, we get $\mathbf{v}' = \{\mathbf{v}'_b,1\leq b\leq|\mathcal{B}|\} = \Psi(\mathcal{B}; \theta')$. Note that the features $\mathbf{v}'$ are different from $\mathbf{v}$, computed in the {\em warm-up} stage, because of the parameter update in Alg.\ref{alg:warm}.
Then, we use the old-task classifiers $\phi_o$ (for evey old tasks $T_o$) to predict the output probabilities using $\mathbf{v}'$. Following a distillation approach, we want that the predictions $\phi_{o}(\mathbf{v}',\tau;\theta')$ should not differ much from the initial values $\hat{\mathbf{y}}^o \in \hat{Y}$. To measure this change in the predictions, we use a cross-entropy loss between the initial and current predicted probability distributions:
\begin{equation}
\label{eq.distillation-loss}
\mathcal{L}_{d} = \sum_{o=1}^{t-1}H(\phi_{o}(\mathbf{v}', \tau;\theta'),\hat{\mathbf{y}}^o) = - \sum_{o=1}^{t-1}\sum_{b=1}^{|\mathcal{B}|} \hat{\mathbf{y}}_b^o  \log \big(\phi_{o}(\mathbf{v}'_b, \tau ; \theta')\big).
\end{equation}
\noindent It is worth noting that the distillation loss is used only in the \emph{joint training} stage and not in the \emph{warm-up} stage. The reason for this choice is that, in the \emph{warm-up} stage, the distillation loss $\mathcal{L}_{d}$ would have a zero gradient since $\phi_{o}(\mathbf{v}'_b,\tau;\theta)=\phi_{o}(\mathbf{v}_b,\tau;\theta)=\hat{\mathbf{y}}_b^o$. Because of the first gradient descent step in the \emph{warm-up} stage, we obtain non-null gradients for the distillation loss in the second stage. This observation mainly motivates our two-stage pipeline.

The distillation loss gradient is weighted using a normalization factor. Specifically, given the distillation gradient $G_j = \frac{\partial \mathcal{L}_{d}}{\partial \theta'}$ and the cross-entropy gradient norm $||G_w||$ computed in the warm-up stage, the gradient is multiplied by $\lambda\frac{||G_w||}{||G_j||}$. The intuition behind this normalization is that we want to balance the two gradients in a dynamic way while training. The parameter $\lambda$ is a static parameter that adjusts the weight of the distillation and the cross-entropy gradients, accounting for the possible imbalance originated with the unconstrained \emph{warm-up} update. 
Finally, we use the new-task classifier $\phi_{t}$ to compute the network predictions and its resulting cross-entropy loss $\mathcal{L}_{t}$. Assuming that the  norm of the gradient of this loss does not change drastically between the two stages (i.e., $||G_w|| \simeq ||\frac{\partial \mathcal{L}_{t}}{\partial \theta'}||$), we can sum $G_j$ with $\frac{\partial \mathcal{L}_{t}}{\partial \theta'}$ (all the gradient terms have a balanced contribution).
    
For  sake of simplicity, we used above the notation $\frac{||G_w||}{||G_j||}$, which includes the gradients of  all the layers of the network. In practice, we actually compute the norms separately for each layer, because in this way the memory cost can be kept extremely small and, empirically, we observed that this leads to a more stable training.

The \emph{joint training} stage can be iterated several times. Empirically, we found iterating twice to be beneficial. Note that, in the second iteration of this stage, $\mathcal{L}_{d}$ and $\mathcal{L}_{t}$ are computed at the value $\theta''$ obtained from the first iteration.

Before proceeding to the next batch, all the memory (including the probability bank $\hat{Y}$) is released, except for the parameters $\theta''$.

\subsection{Memory Efficient Data Augmentation}
Data augmentation is a widely-used technique in CL. However, in the extreme memory constraint scenario, standard data augmentation procedures cannot be used since it would result in an important memory cost. We propose a specific data-augmentation procedure that is integrated in our \textit{BLD} framework. We use a set of data augmentation techniques (e.g., image cropping, flip, rotation, color jittering etc.) in order to artificially populate $\mathcal{B}$. In the warm-up stage, when filling the probability bank $\hat{Y}$, we also store the transformation type (e.g. rotation) and possible parameters (e.g. angle). However, we do not store the transformed images. Consequently, the memory cost of data augmentation remains negligible with respect to the batch and network memory sizes. In the \emph{joint training} stage, when computing the feature $\mathbf{v}'$, we read the transformation information stored together with the probability bank and use it to re-generate the transformed images. The transformed image is then provided as input to the feature extractor. In this way, we use the same data augmentation in the two stages without requiring to store all the augmented images.

\section{Experiments}

\subsection{Experimental Protocol}

\textbf{Datasets.} We measure the performance of the proposed solution to MC-OCL using accuracy on three publicly available and widely used datasets.

The {MNIST} \cite{lecunmnist} and {SVHN} \cite{netzer2011reading} datasets are composed of images depicting digits. In our experiments, 
both datasets are split into different tasks with non-overlapping classes. We choose not to perform experiments on the permuted variant of MNIST, since it has been shown to be a poor benchmark for CL \cite{farquhar2018towards}. Some previous works \cite{aljundi2019online,aljundi2019task,lopez2017gradient} prefer to extract a small subset of the samples for training. Instead, consistently with the most prior art, we use all the training data available. This choice enables us to assess which methods are robust to a large number of gradient steps and which are not.

CIFAR10 \cite{cifar} is also split into disjoint tasks as in {\cite{aljundi2019gradient,aljundi2019online}}, with the difference that, given the memory constraints we introduce, we cannot store any data and therefore we are unable to perform validation. Consequently, we use all the training samples for training. 

For all datasets we split the data into 2 and 5 tasks, which generates subsets of 5 and 2 classes respectively. This enables finer behavioral analysis of the model with respect to short and long task sequences. The splits are performed randomly, but, for fairness, we run all methods on the same splits to minimize the bias that different splits could introduce in the evaluation.

\textbf{Implementation details.}
Throughout all experiments, regardless of the dataset and the number of tasks, we employ a ResNet18 \cite{he2016deep} as a feature extractor. As per Sec. \ref{sec:method}, on top of the feature extractor we use a classifier composed of a linear layer and a softmax. As soon as a new task starts, a new classifier is instantiated with randomly initialized weights and biases.

For all experiments that only require a single sweep through the data we train on batches composed of 20 images, randomly sampled from a task-specific subset of data. We found this batch size to be the right trade-off for our experiments, since it well approximates the online setup without preventing the model from properly learning new tasks. These batches are then transformed 50 times and forwarded into the network. The gradients generated by all losses are averaged over these transformations. Note that these operations do not require any additional memory, since the transformations can be applied right before the forward pass, without storing the augmented images, and gradients can be averaged in-place. For the details on this matter refer to Algorithm \ref{alg:warm} and \ref{alg:joint}. In Pytorch \cite{paszke2017automatic} this can be implemented by calling \texttt{backward()} multiple times (one for each transformation) without performing any optimization steps in-between. 

Two iterations are performed for every joint training stage, with learning rate {$\alpha_j$ set to $10^{-4}$, while the \emph{warm-up} stage is performed only once with a learning rate $\alpha_w = 10^{-2} \cdot \alpha_j$. The parameter $\lambda$ has a value of 2.
For {\em offline LwF}} \cite{li2017learning}, instead, batches contain 500 images each and only one transformation is computed per batch. Depending on the dataset and the number of tasks we train the model for a different number of epochs, ranging from 10 epochs for MNIST (5-tasks) to 120 epochs for CIFAR10 (2-task) {with learning rate equal to $10^{-4}$.} For all the methods, we run each experiment 5 times and report the average accuracy. Note that our  method (\textit{BLD}) is trained using only one epoch.

\subsection{Experimental evaluation}
\label{sec:baseline}
\textbf{Baselines.} {Our method} can be accommodated among regularization-based methods, which in turn can be divided into prior-based and data-driven categories. However, we do not consider prior-based baselines such as EWC \cite{kirkpatrick2017overcoming} as they have been shown to work poorly in the online setting \cite{aljundi2019online}, {and do not satisfy the MC-OCL constraints.} Instead, we include an extensive comparison with {\emph{LwF}} \cite{li2017learning}, which is the most similar data-driven method to ours. Therefore, we consider the following reference baselines:
\begin{itemize}\renewcommand{\labelitemi}{$\bullet$} 
\item \textit{Finetune.} It trains continuously as the data for the new task is available without any attempt to avoid forgetting;
\item  \textit{Batch-level L2}, denoted as \emph{L2}, is a na\"ive baseline we devised specifically for CL with extreme memory constraints. For every incoming batch it saves a copy of the parameters before the model gets updated. Subsequently, it proceeds to update the network, first with a \emph{warm-up} step, similar to the \emph{warm-up} we perform for our method, and then with a series of joint steps. These joint steps are the result of the back-propagation of two losses: the cross-entropy loss with respect to the current task and the \emph{L2} loss between current and previous parameters. 
\item  \textit{Offline LwF} \cite{li2017learning} (upper-bound) trained using multiple passes through the data, sampled i.i.d.. We use a variable number of epochs, depending on the size and the complexity of the dataset, while the batch size is fixed.
\item  \textit{Single-pass LwF} \cite{li2017learning} is a modified version of \emph{LwF}, in which only a single-pass through the data is performed. The distillation mechanism is implemented as in the original offline version. Note that, although each sample is only processed once, this variant can not be considered fully online because it still needs to compute the predictions for the whole task beforehand.
\end{itemize}

\begin{table}[!t]
\scriptsize
\setlength{\tabcolsep}{5pt}

\caption{\label{tab:results_2_tasks}Final test accuracy on MNIST, CIFAR10 and SVHN with 2 tasks}
\begin{center}
    \begin{tabular}{ccccccccccc}
        \toprule
        \multicolumn{2}{c}{\multirow{2.4}{*}{\textbf{Method}}}
        &\multicolumn{3}{c}{\textbf{MNIST}}&\multicolumn{3}{c}{\textbf{CIFAR10}}&\multicolumn{3}{c}{\textbf{SVHN}} \\
        \cmidrule(lr){3-5}
        \cmidrule(lr){6-8}
        \cmidrule(lr){9-11}
        &&T0 & T1 & Avg. &  T0  & T1 & Avg. & T0  & T1 & Avg. \\
        \midrule
        \multirow{3}{*}{MC-OCL}
        &\emph{Finetune} & 80.8 & 99.6 & 90.2 & 60.4 & 85.6 & 73.0 & 78.9 & 95.5 & 87.2 \\
        &\emph{L2}  & 91.7 & 99.6 & \textbf{95.7} & 70.7 & 84.0 & 77.4 & 82.8 & 96.2 & 89.5 \\
        &\emph{BLD} & 89.6 & 99.5 & 94.5 & 70.0 & 86.0 & \textbf{78.0} & 88.2 & 96.2 & \textbf{92.2} \\
        \midrule
        \multirow{1}{*}{Single-pass}
        &\emph{LwF} & 98.2 & 99.7 & 98.9 & 75.7 & 85.8 & 80.7 & 91.5 & 95.6 & 93.5 \\
        \midrule
        \multirow{1}{*}{Offline}
        &\emph{LwF} & 99.5 & 99.8 & 99.7 & 89.6 & 93.0 & 91.3 & 93.9 & 96.3 & 95.1 \\
        \bottomrule
    \end{tabular}
\end{center}
\end{table}

\noindent\textbf{Results and Analysis.}
Tab. \ref{tab:results_2_tasks} shows the performance of the evaluated methods on MNIST, CIFAR10 and SVHN on the 2-task scenario. 
Looking at the results of the \emph{Finetune} model, the difference in performance between the two tasks {T0} and {T1} shows that the \emph{Finetune} model suffers from catastrophic forgetting. The difference is especially important in the case of the MNIST ($18.8\%$) and CIFAR10 ($25.2\%$) datasets. We observe that \emph{L2} mitigates this catastrophic forgetting issue reaching a higher average accuracy in the three datasets, at the cost of a higher memory consumption. \emph{BLD} consistently improves the performance over all the datasets. Our method, obtains better scores on the task T0 compared to the \emph{Finetune} baseline. For CIFAR10 and SVHN, we {also} observe that \emph{BLD} outperforms \emph{Finetune} on T1, possibly due to the fact that some information from T0 has been used to improve  {the} performance on T1 (\emph{forward-transfer}). Overall, \emph{BLD} reaches the best performance in two datasets out of three. Only \emph{L2} performs slightly better on MNIST but requiring much more memory. 

When it comes to comparing to the offline baseline that can have access to each image several times, we observe that our method can bridge half of the gap between \emph{Finetune} and the offline \emph{LwF} on the MNIST and SVHN datasets. Interestingly, \textit{BLD} is able to obtain results close to the \emph{single-pass LwF} on the SVHN dataset even though the latter breaks constraint (1) of MC-OCL.

Concerning the  {5-tasks experiments}, results are reported in Tab. \ref{tab:resultsMNIST}, \ref{tab:resultsCIFAR} and \ref{tab:resultsSVHN} for the MNIST, CIFAR10 and SVHN datasets, respectively. Note that, for every method, we also report the memory overhead. More specifically, we report the memory storage that is required by every method while training on the current batch (\emph{Intra-batch}), when switching between batches (\emph{Inter-batch}) and for data storage in the case of non-online methods. We report memory in bytes.

\begin{table}[t]
\scriptsize
\setlength{\tabcolsep}{2.8pt}
\caption{\label{tab:resultsMNIST}Final test accuracy on MNIST with 5 tasks}
\begin{center}
    \begin{tabular}{ccccccccccc}
        \toprule
        \multicolumn{2}{c}{\multirow{2.4}{*}{\textbf{Method}}}
        &\multicolumn{6}{c}{\textbf{MNIST}}&\multicolumn{3}{c}{\textbf{Memory Overhead}} \\
        \cmidrule(lr){3-8}
        \cmidrule(lr){9-11}
        && T0 & T1 & T2 & T3 & T4 & Avg. & Intra-batch & Inter-batch & Data Storage \\ 
        \midrule
        \multirow{3}{*}{MC-OCL}
        &\emph{Finetune} & 66.6 & 68.0 & 76.8 & 91.8 & 99.8 & 80.6 & - & - & -\\
        &\emph{L2} & 54.9 & 55.7 & 85.7 & 94.0 & 99.8 & 78.0 & 44.8MB & - & -\\
        &\emph{BLD} & 78.0 & 82.5 & 93.0 & 96.4 & 99.7 & \textbf{89.9} & 32kB & - & -\\
        \midrule
        \multirow{1}{*}{Single-pass}
        & $LwF$ & 98.2 & 99.4 & 98.5 & 99.8 & 99.8 & 99.1 & 384kB & 384kB & 2MB\\
        \midrule
        \multirow{1}{*}{Offline}
        &$LwF$ & 99.5 & 99.6 & 98.0 & 99.8 & 99.8 & 99.3 & 384kB & 384kB & 2MB\\
        \bottomrule
    \end{tabular}
\end{center}
\end{table}

\begin{table}[t]
\scriptsize
\setlength{\tabcolsep}{2.8pt}
\caption{\label{tab:resultsCIFAR}Final test accuracy on CIFAR10 with 5 tasks}
\begin{center}
    \begin{tabular}{ccccccccccc}
        \toprule
        \multicolumn{2}{c}{\multirow{2.4}{*}{\textbf{Method}}}
        &\multicolumn{6}{c}{\textbf{CIFAR10}}&\multicolumn{3}{c}{\textbf{Memory Overhead}} \\
        \cmidrule(lr){3-8}
        \cmidrule(lr){9-11}
        && T0 & T1 & T2 & T3 & T4 & Avg. & Intra-batch & Inter-batch & Data Storage \\ 
        \midrule
        \multirow{3}{*}{MC-OCL}
        &\emph{Finetune} & 59.6 & 58.2 & 66.8 & 80.2 & 97.0 & 72.3 & - & - & - \\
        &\emph{L2} & 75.5 & 65.3 & 73.5 & 81.3 & 96.8 & 78.5 & 44.8MB & - & -\\
        &\emph{BLD} & 83.4 & 83.2 & 79.5 & 88.1 & 97.0 & \textbf{86.2} & 32kB & - & - \\
        \midrule
        \multirow{1}{*}{Single-pass}
        &\emph{LwF}  & 81.2 & 83.6 & 81.1 & 88.5 & 96.5 & 86.2 & 320kB & 320kB & 36.8MB \\
        \midrule
        \multirow{1}{*}{Offline}
        &\emph{LwF} & 93.8 & 94.1 & 91.6 & 96.2 & 98.3 & 94.8 & 320kB & 320kB & 36.8MB\\
        \bottomrule
    \end{tabular}
\end{center}
\end{table}

\begin{table}[t]
\scriptsize
\setlength{\tabcolsep}{2.8pt}
\caption{\label{tab:resultsSVHN}Final test accuracy on SVHN with 5 tasks}
\begin{center}
    \begin{tabular}{ccccccccccc}
        \toprule
        \multicolumn{2}{c}{\multirow{2.4}{*}{\textbf{Method}}}
        &\multicolumn{6}{c}{\textbf{SVHN}}&\multicolumn{3}{c}{\textbf{Memory Overhead}} \\
        \cmidrule(lr){3-8}
        \cmidrule(lr){9-11}
        && T0 & T1 & T2 & T3 & T4 & Avg. & Intra-batch & Inter-batch & Data Storage \\ 
        \midrule
        \multirow{3}{*}{MC-OCL}
        &\emph{Finetune} & 65.9 & 60.6 & 77.5 & 87.6 & 98.4 & 78.0 & - & - & -\\
        &\emph{L2}  & 75.2 & 61.8 & 90.9 & 93.4 & 98.1 & 81.3 & 44.8MB & -& -\\
        &\emph{BLD}  & 78.5 & 79.6 & 92.1 & 95.7 & 98.1 & \textbf{88.8} & 32kB & - & -\\
        \midrule
        \multirow{1}{*}{Single-pass}
        &\emph{LwF}  & 78.9 & 91.5 & 94.3 & 95.6 & 98.2 & 91.7 & 469kB & 469kB & 47.2MB \\
        \midrule
        \multirow{1}{*}{Offline}
        &\emph{LwF}  & 97.7 & 97.8 & 97.2 & 98.7 & 98.9 & 98.1 & 469kB & 469kB & 47.2MB\\
        \bottomrule
    \end{tabular}
\end{center}
\end{table}

In the three datasets, we again observe strong catastrophic forgetting in the case of the \emph{Finetune} model.  {Again,} \emph{L2} prevents forgetting to some extent but it has a high intra-batch memory overhead since it requires to store a copy of the network parameters. Despite its lower memory overhead, our approach reaches the best performance on the three datasets with a significant margin with respect to \emph{L2} ($+11.9\%$, $7.7\%$ and $7.5\%$, respectively). This result is extremely interesting since it shows that \textit{BLD} can prevent the network from drifting and forgetting even for longer sequences of tasks. 

When it comes to offline methods, they both outperform our proposed method. Nevertheless, we observe that \textit{BLD} reaches the same performance as \emph{Single-pass} \textit{LwF} on CIFAR10, which requires access to the complete training set of the current task. This requirement leads non-negligible data storage depending on the dataset (from 2MB to 47.MB for SVHN). Note that, the data storage requirement grows linearly with the size of the dataset. In addition, we observe that both methods require an intra-batch memory overhead approximately ten times higher than our approach.

\subsubsection{Ablation Study.}
\begin{table}[t]
\scriptsize
\setlength{\tabcolsep}{3.3pt}
\caption{\label{tab:ablation} Ablation Study on the CIFAR10 datatset with 2 and 5 tasks}
  \begin{center}
\subfloat[text][2 tasks (5 classes each)]{
    \begin{tabular}{cccc}
        \toprule
        \textbf{Method} & \textbf{T0} & \textbf{T1} & \textbf{Avg.} \\ 
        \midrule
        \emph{Finetune} & 60.4 & 85.6 & 73.0 \\
        \emph{L2} & 70.7 & 84.0 & 77.4 \\
        \textit{Alternated}  & 57.8 & 85.8 & 71.8 \\
        \textit{No-balancing} & 61.4 & 86.3 & 73.8 \\
        \textit{Full} & 70.0 & 86.0 & \textbf{78.0} \\
        \bottomrule
\end{tabular}}\hspace{1cm}
\subfloat[text][5 tasks (2 classes each)]{
  \begin{tabular}{ccccccc}
        \toprule
        \textbf{Method} & \textbf{T0} & \textbf{T1} & \textbf{T2} & \textbf{T3} & \textbf{T4} & \textbf{AVG} \\ 
        \midrule
        \emph{Finetune} & 59.6 & 58.2 & 66.8 & 80.2 & 97.0 & 72.3 \\
        \emph{L2} & 75.5 & 65.3 & 73.5 & 81.3 & 96.8 & 78.5 \\
        \textit{Alternated}  & 77.7 & 74.5 & 70.2 & 87.1 & 96.9 & 81.3 \\
        \textit{No-balancing} & 78.5 & 72.9 & 74.9 & 85.0 & 96.9 & 81.6 \\
        \textit{Full} & 83.4 & 83.2 & 79.5 & 88.1 & 97.0 & \textbf{86.2} \\
         \bottomrule
    \end{tabular}
}
  \end{center}
\end{table}

 {We perform} an in-depth ablation study to evaluate each component of the proposed method. In addition to the \textit{Finetune} and the \textit{L2} baselines described above, we compare with the following variants of our models:
\textit{Alternated}, a model that does not perform joint updates but simply alternates between a learning step on the new task and a distillation step, and \textit{No-balancing}, a variant of our model that uses our two-stage approach but where the cross-entropy $\mathcal{L}_{t}$ and distillation $\mathcal{L}_{d}$ losses are not dynamically balanced. More precisely, this method is equivalent to our full model replacing $\frac{||G_w||}{||G_j||}$  {with} 1 in Algorithm \ref{alg:joint}.  {In Tab.~\ref{tab:ablation},} \textit{Full} denotes the full model as described in Sec.~\ref{Setting}.

The results of the ablation study are reported in Tab.\ref{tab:ablation}. As previously observed, the \emph{Finetune} model suffers from catastrophic forgetting. The forgetting problem is even clearer on the 5-task setting.  As in previous experiments, \emph{L2} helps preventing forgetting but breaks our  {proposed} constraints. \emph{Alternated} improves the performance on the 5-task setting but deteriorates on the 2-task setting, showing that naively alternating between new task learning and distillation is not enough in our challenging scenario. Conversely, we observe that the \emph{No-balancing} model improves the performance with respect to \emph{Finetune} in both settings. Note that, in the 2-task setting, \emph{No-balancing} and the full model outperform \emph{Finetune} on T1. This shows that our two-stage pipeline might produce some \textit{forward transfer} from task T0 to T1. On the 5-task setting, the gain of \emph{No-balancing} is more important ($+9.3\%$ with respect to \emph{Finetune} and $+0.9\%$ with respect to \emph{alternated}). Finally, using our dynamic gradient weighting with balancing leads to further improvement reaching the highest performance. The gain in performance is consistent over all the tasks and is especially clear for the first tasks. This ablation study experimentally confirms the importance of the two-stage approach and the dynamic gradient weighting.



\section{Conclusions}

In this paper we proposed setting that allows us to study continual learning under extreme memory constraints. More precisely, we impose two constraints: 1) No information is passed between batches and tasks; 2) No auxiliary network can be used. To tackle this setting that cannot be addressed by the current methods, we introduced Batch-level Distillation. Based on knowledge distillation, our approach proceeds in two stages where, first, we start learning the new task classifier and compute old classifier predictions, and then, we perform a joint training using both distillation and the new task loss. We evaluated our method on three publicly available datasets and show that \textit{BLD} can efficiently prevent catastrophic forgetting. As future work, we plan to extend \textit{BLD} to other problems such as image segmentation and object detection.

\section*{Acknowledgements}
We acknowledge financial support from the European Institute of Innovation \& Technology (EIT) and the H2020 EU project SPRING - Socially Pertinent Robots in Gerontological Healthcare. This work was carried out under the ``Vision and Learning joint Laboratory” between FBK and UNITN.

\bibliographystyle{splncs04}
\bibliography{egbib}

\begin{thebibliography}{10}
\providecommand{\url}[1]{\texttt{#1}}
\providecommand{\urlprefix}{URL }
\providecommand{\doi}[1]{https://doi.org/#1}

\bibitem{aljundi2018memory}
Aljundi, R., Babiloni, F., Elhoseiny, M., Rohrbach, M., Tuytelaars, T.: Memory
  aware synapses: Learning what (not) to forget. In: ECCV (2018)

\bibitem{aljundi2019online}
Aljundi, R., Belilovsky, E., Tuytelaars, T., Charlin, L., Caccia, M., Lin, M.,
  Page-Caccia, L.: Online continual learning with maximal interfered retrieval.
  In: NeurIPS. pp. 11849--11860 (2019)

\bibitem{aljundi2019task}
Aljundi, R., Kelchtermans, K., Tuytelaars, T.: Task-free continual learning.
  In: CVPR (2019)

\bibitem{aljundi2019gradient}
Aljundi, R., Lin, M., Goujaud, B., Bengio, Y.: Gradient based sample selection
  for online continual learning. In: NeurIPS (2019)

\bibitem{castro2018end}
Castro, F.M., Mar{\'\i}n-Jim{\'e}nez, M.J., Guil, N., Schmid, C., Alahari, K.:
  End-to-end incremental learning. In: ECCV (2018)

\bibitem{cermelli2020modeling}
Cermelli, F., Mancini, M., Bulo, S.R., Ricci, E., Caputo, B.: Modeling the
  background for incremental learning in semantic segmentation. In: CVPR. pp.
  9233--9242 (2020)

\bibitem{chaudhry2018riemannian}
Chaudhry, A., Dokania, P.K., Ajanthan, T., Torr, P.H.: Riemannian walk for
  incremental learning: Understanding forgetting and intransigence. In: ECCV
  (2018)

\bibitem{dhar2019learning}
Dhar, P., Singh, R.V., Peng, K.C., Wu, Z., Chellappa, R.: Learning without
  memorizing. In: CVPR (2019)

\bibitem{Duda2000}
Duda, R.O., Hart, P.E., Stork, D.G.: Pattern Classification (2nd Edition).
  Wiley-Interscience (2000)

\bibitem{farquhar2018towards}
Farquhar, S., Gal, Y.: Towards robust evaluations of continual learning. arXiv
  preprint arXiv:1805.09733  (2018)

\bibitem{he2016deep}
He, K., Zhang, X., Ren, S., Sun, J.: Deep residual learning for image
  recognition. In: CVPR. pp. 770--778 (2016)

\bibitem{hinton2015distilling}
Hinton, G., Vinyals, O., Dean, J.: Distilling the knowledge in a neural
  network. stat  (2015)

\bibitem{hou2019learning}
Hou, S., Pan, X., Loy, C.C., Wang, Z., Lin, D.: Learning a unified classifier
  incrementally via rebalancing. In: CVPR (2019)

\bibitem{javed2019meta}
Javed, K., White, M.: Meta-learning representations for continual learning. In:
  NeurIPS. pp. 1820--1830 (2019)

\bibitem{kirkpatrick2017overcoming}
Kirkpatrick, J., Pascanu, R., Rabinowitz, N., Veness, J., Desjardins, G., Rusu,
  A.A., Milan, K., Quan, J., Ramalho, T., Grabska-Barwinska, A., et~al.:
  Overcoming catastrophic forgetting in neural networks. PNAS  (2017)

\bibitem{cifar}
Krizhevsky, A.: Learning multiple layers of features from tiny images. Tech.
  rep. (2009)

\bibitem{DeLange-CL-survey}
Lange, M.D., Aljundi, R., Masana, M., Parisot, S., Jia, X., Leonardis, A.,
  Slabaugh, G.G., Tuytelaars, T.: Continual learning: {A} comparative study on
  how to defy forgetting in classification tasks. arXiv:1909.08383  (2019)

\bibitem{lecunmnist}
LeCun, Y., Cortes, C.: {MNIST} handwritten digit database  (2010),
  \url{http://yann.lecun.com/exdb/mnist/}

\bibitem{lee2020neural}
Lee, S., Ha, J., Zhang, D., Kim, G.: A neural dirichlet process mixture model
  for task-free continual learning. In: ICLR (2020)

\bibitem{li2017learning}
Li, Z., Hoiem, D.: Learning without forgetting. IEEE T-PAMI  (2017)

\bibitem{lopez2017gradient}
Lopez-Paz, D., Ranzato, M.: Gradient episodic memory for continual learning.
  In: NIPS (2017)

\bibitem{mallya2018packnet}
Mallya, A., Lazebnik, S.: Packnet: Adding multiple tasks to a single network by
  iterative pruning. In: CVPR (2018)

\bibitem{netzer2011reading}
Netzer, Y., Wang, T., Coates, A., Bissacco, A., Wu, B., Ng, A.Y.: Reading
  digits in natural images with unsupervised feature learning  (2011)

\bibitem{ostapenko2019learning}
Ostapenko, O., Puscas, M., Klein, T., Jahnichen, P., Nabi, M.: Learning to
  remember: A synaptic plasticity driven framework for continual learning. In:
  CVPR (2019)

\bibitem{paszke2017automatic}
Paszke, A., Gross, S., Chintala, S., Chanan, G., Yang, E., DeVito, Z., Lin, Z.,
  Desmaison, A., Antiga, L., Lerer, A.: Automatic differentiation in pytorch
  (2017)

\bibitem{rebuffi2017icarl}
Rebuffi, S.A., Kolesnikov, A., Sperl, G., Lampert, C.H.: icarl: Incremental
  classifier and representation learning. In: CVPR (2017)

\bibitem{riemer2018learning}
Riemer, M., Cases, I., Ajemian, R., Liu, M., Rish, I., Tu, Y., Tesauro, G.:
  Learning to learn without forgetting by maximizing transfer and minimizing
  interference. arXiv preprint arXiv:1810.11910  (2018)

\bibitem{rusu2016progressive}
Rusu, A.A., Rabinowitz, N.C., Desjardins, G., Soyer, H., Kirkpatrick, J.,
  Kavukcuoglu, K., Pascanu, R., Hadsell, R.: Progressive neural networks. arXiv
  preprint arXiv:1606.04671  (2016)

\bibitem{shin2017continual}
Shin, H., Lee, J.K., Kim, J., Kim, J.: Continual learning with deep generative
  replay. In: NeurIPS (2017)

\bibitem{wu2018memory}
Wu, C., Herranz, L., Liu, X., van~de Weijer, J., Raducanu, B., et~al.: Memory
  replay gans: Learning to generate new categories without forgetting. In:
  NeurIPS (2018)

\bibitem{wu2019large}
Wu, Y., Chen, Y., Wang, L., Ye, Y., Liu, Z., Guo, Y., Fu, Y.: Large scale
  incremental learning. In: CVPR (2019)

\end{thebibliography}
\end{document}